\documentclass[lettersize,journal]{IEEEtran}
\usepackage{amsmath,amsfonts,amssymb, amsthm}
\usepackage{algorithmic}
\usepackage{algorithm}
\usepackage{caption}
\usepackage{array}
\usepackage[caption=false,font=normalsize,labelfont=sf,textfont=sf]{subfig}
\usepackage{textcomp}
\usepackage{stfloats}
\usepackage{url}
\usepackage{verbatim}
\usepackage{graphicx}
\usepackage{cite}
\usepackage[numbers,sort&compress]{natbib}

\usepackage{tabularx}
\usepackage{float}
\usepackage{mathtools}
\usepackage{booktabs}
\usepackage{multirow}
\usepackage{amsmath}
\usepackage{amssymb}

\def\BibTeX{{\rm B\kern-.05em{\sc i\kern-.025em b}\kern-.08em
    T\kern-.1667em\lower.7ex\hbox{E}\kern-.125emX}}
\usepackage{balance}

\newcommand{\figref}[1]{Fig.~\ref{#1}}
\newcommand{\tabref}[1]{Tab.~\ref{#1}}
\newcommand{\secref}[1]{Sec.~\ref{#1}}

\newcommand{\equref}[1]{Eq.~\ref{#1}}
\newcommand{\thmref}[1]{Theorem~\ref{#1}}
\newcommand{\defref}[1]{Definition~\ref{#1}}
\newcommand{\lemref}[1]{Lemma~\ref{#1}}

\usepackage{hyperref}
\hypersetup{
    colorlinks=true,
    linkcolor=blue,
    filecolor=blue,      
    urlcolor=blue,
    citecolor=cyan,
}

\newtheorem{definition}{Definition}
\newtheorem{lemma}{Lemma}

\newtheorem{theorem}{Theorem}

\newcommand\highlightReference[1]{%
  \expandafter\newcommand\csname highlightReference-#1\endcsname{}%
}
\let\oldbibitem\bibitem
\def\bibitem#1 #2\par{%
  \expandafter\ifx\csname highlightReference-#1\endcsname\relax
    \oldbibitem{#1}#2\par
  \else
    \oldbibitem{#1}\highlight{#2}\par
  \fi
}
\usepackage{color}
\newcommand\highlight[1]{\textcolor{red}{#1}}

\begin{document}
\bstctlcite{IEEEexample:BSTcontrol}

\title{High-speed and High-quality Vision Reconstruction of Spike Camera with Spike Stability Theorem}
\author{
  \IEEEauthorblockN{
    Wei Zhang, %
    Weiquan Yan, %
    Yun Zhao, %
    Wenxiang Cheng,
    Gang Chen,
    Huihui Zhou, and %
    Yonghong Tian \thanks{This work was supported by the National Natural Science Foundation of China under Grant 62027804 and the Major Key Project of PCL (PCL2021A13).  \\ \\ Wei Zhang, Weiquan Yan, Yun Zhao, Wenxiang Chen, Gang Chen, Huihui Zhou, and Yonghong Tian are with the Department of Networked Intelligence, Pengcheng Laboratory, Shenzhen 518000, China. Gang Chen is also with the School of Computer Science and Engineering, Sun Yat-sen University, Guangzhou, China. Yonghong Tian is also with the School of Computer Science, Peking University, Beijing 100871, China(E-mail: zhangwei1213052@126.com, ywqqqqqq@outlook.com, zhaoy@pcl.ac.cn, chengwx01@pcl.ac.cn, cheng83@mail.sysu.edu.cn, zhouhh@pcl.ac.cn, yhtian@pku.edu.cn)}%
  }
}

\maketitle
\begin{abstract}
Neuromorphic vision sensors, such as the dynamic vision sensor (DVS) and spike camera, have gained increasing attention in recent years. The spike camera can detect fine textures by mimicking the fovea in the human visual system, and output a high-frequency spike stream. Real-time high-quality vision reconstruction from the spike stream can build a bridge to high-level vision task applications of the spike camera. To realize high-speed and high-quality vision reconstruction of the spike camera, we propose a new spike stability theorem that reveals the relationship between spike stream characteristics and stable light intensity. Based on the spike stability theorem, two parameter-free algorithms are designed for the real-time vision reconstruction of the spike camera. To demonstrate the performances of our algorithms, two datasets (a public dataset PKU-Spike-High-Speed and a newly constructed dataset SpikeCityPCL) are used to compare the reconstruction quality and speed of various reconstruction methods. Experimental results show that, compared with the current state-of-the-art (SOTA) reconstruction methods, our reconstruction methods obtain the best tradeoff between the reconstruction quality and speed. Additionally, we design the FPGA implementation method of our algorithms to realize the real-time (running at 20,000 FPS) visual reconstruction. Our work provides new theorem and algorithm foundations for the real-time edge-end vision processing of the spike camera.
\end{abstract}

\begin{IEEEkeywords}
Spike stability theorem, spike camera, high-speed reconstruction, FPGA.
\end{IEEEkeywords}

\section{Introduction}
\label{sec:intro}

\IEEEPARstart{H}{uman} vision system is powerful but complicated. Approaching or even outperforming humans in vision tasks is the ultimate goal of the machine vision system. The traditional camera does not work properly in extreme cases, e.g., extremely dark or bright environments, and high-speed moving environments. Therefore, bio-inspired sensors attract more and more attention in unmanned aerial vehicles \cite{salt2019parameter}, autonomous vehicles \cite{chen2020event}, humanoid robots \cite{glover2017robust}, etc., due to their advantages of high dynamic range and high temporal resolution \cite{lichtsteiner2008128, posch2010qvga, brandli2014240, lenero2013bio, guo2017live, ding2023mlb, xu2023denoising, nie2020high, huang20221000, zhu2020hybrid}.

Intuitively, bio-inspired sensors simulate the human visual system. Retina, the most important part of the eye, receives incoming photons and converts them to binary electrical signals \cite{luo2020principles}. Specifically, there are three different ganglion cells: M-type ganglion cell, P-type ganglion cell, and nonM-nonP ganglion cell \cite{bear2020neuroscience}. The M-type ganglion cell is sensitive to the change of light intensity and thus sensitive to moving objects. The P-type ganglion cell, nevertheless, perceives the fine texture and contour of the surrounding environment. Some of the nonM-nonP ganglion cells are sensitive to the wavelength of light. In recent years, sensors mimicking M-type and P-type ganglion cells are proposed named DVS (corresponding to M-type cells) \cite{lichtsteiner2008128, posch2010qvga, lenero2013bio, brandli2014240, guo2017live, ding2023mlb, xu2023denoising, nie2020high} and spike camera (corresponding to P-type cells) \cite{huang20221000, zhu2020hybrid}. The DVS can detect the change of light intensity efficiently which is suitable for capturing the motion part of the scene. Different from the DVS, the spike camera can perceive the fine texture of the surrounding environment. Concretely, the DVS adopts a differential circuit to capture every micro change of light intensity and emits spikes asynchronously. In contrast, the spike camera integrally accumulates the energy of photons and emits spikes once the integrator (capacitance) is fully charged.  

The majority of high-level vision tasks (such as classification, object detection, and depth estimation) highly rely on the texture and structure information which is hard to extract from spike stream directly. So, a well-reconstructed image can provide a foundation for these tasks. Many researchers focus on reconstructing texture using DVS. Earlier reconstruction algorithms depend on hand-crafted features like temporal filtering\cite{scheerlinck2018continuous, scheerlinck2019asynchronous} and regularization \cite{bardow2016simultaneous, munda2018real}. Recent works adopt data-driven methods to obtain higher-quality reconstructed images \cite{paredes2021back, rebecq2019events, rebecq2019high, ronneberger2015u, scheerlinck2020fast, stoffregen2020reducing, wang2019event, weng2021event}. However, the spike sampling law of DVS determines the information loss of fine texture of the relatively static object. To solve this problem, some methods that add the information from the frame-based image sensor \cite{pan2019bringing, zhou2023deblurring, chen2022residual, jiang2023event} or the spike camera \cite{zhu2021neuspike} were proposed to assist the texture reconstruction of DVS. These methods can obtain better texture reconstruction and keep the advantage of a high frame rate.   

Different from DVS, the spike camera will emit the spike stream even if the scene is static. This special feature attracts researchers to study different vision tasks of the spike camera, such as compression coding \cite{zhu2020hybrid}, denoising \cite{xu2021denoising}, object recognition \cite{zhao2023spireco}, depth estimation \cite{zhang2022ultra}, vision reconstruction\cite{zhu2019retina, zhu2020retina, zhao2020high, zheng2021high, zhao2021spk2imgnet, zhao2022reconstructing, chen2022self, chen2023self, zhu2023ultra, zheng2023capture}, and super-resolution \cite{xiang2023learning, zhao2021super}. Similar to the DVS, the vision reconstruction of the spike camera is important as it can provide a foundation for different high-level vision tasks. Based on the physical rule of spike emission, basic physical-model-based reconstruction methods, texture from the inter-spike interval (TFI) and texture from playback (TFP), were proposed for visual reconstruction \cite{zhu2019retina}. Besides, different neural-networks-model-based methods \cite{zhu2020retina, zhao2020high, zheng2021high, zhao2021super, zhao2021spk2imgnet, zhao2022reconstructing, chen2022self, xiang2023learning, chen2023self, zhu2023ultra, zheng2023capture} were also reported to reconstruct the texture. Despite many reconstruction methods having emerged in the last few years, only TFI and TFP methods were reported for the real-time application of the spike camera \cite{zhang2022ultra}. Unfortunately, TFI and TFP suffer from motion blur and noise, resulting in unsatisfactory reconstruction images. Compared with TFI and TFP, the neural-networks-model-based methods can reconstruct higher-quality images, but they are commonly complex and unsuitable for real-time edge computing platform, which loses the high sampling rate (the sampling rate of a portable spike camera is 20,000Hz \cite{zhu2021neuspike, xu2021denoising}) benefit of the spike camera. Consequently, how to realize high-speed and high-quality vision reconstruction is still a question for the spike camera.

In this paper, we propose and prove a new spike stability theorem for the visual reconstruction of the spike camera. The spike stability theorem illustrates how the spike stream will be output under a certain light intensity. Based on the spike stability theorem, we further develop two parameter-free algorithms for high-quality and high-speed vision reconstruction of the spike camera and design their FPGA deployment process. We experimentally compare our algorithms with existing reconstruction algorithms, and the results show that our methods obtain the best tradeoff between reconstruction quality and speed. We also accelerate successfully our methods to 20,000 FPS (frame per second) by the FPGA platform, realizing the real-time high-quality vision reconstruction of the spike camera. The key contributions of our work are as follows:
\begin{itemize}
\item[1)] We propose a new spike stability theorem, mathematically revealing the relationship between spike stream characteristics and stable light intensity. This theorem not only lays a foundation for vision reconstruction of the spike camera but also for the future design of high-level vision task algorithms.  
\item[2)] We design two parameter-free algorithms for the vision reconstruction of the spike camera based on the new spike stability theorem. The two algorithms solve the motion blur and noise problem of physical-model-based reconstruction methods and the speed problem of neural-network-model-based reconstruction methods, achieving the best tradeoff between the reconstruction quality and speed.
\item[3)] We design the FPGA implementation method of our algorithms and achieve a reconstruction speed of 20,000 FPS, providing a foundation for the real-time edge-end high-level vision processing of the spike camera.
\item[4)] We provide a new spike dataset named SpikeCityPCL. The SpikeCityPCL can be used for the vision reconstruction method investigation of the spike camera.
\end{itemize}

\section{Related Works}
\label{sec:relate}

\textbf{Spike camera:} Spike camera is a fovea-like sensor mimicking human vision\cite{huang20221000}. It perceives the texture information following the neurological integral-and-fire rule. Spike camera consists of three basic components: photodiode, trigger and reset circuit, and readout circuit. The photodiode first captures the incoming photons and converts them to the photocurrent. The voltage $V$ of the photodiode increases with the accumulation of photons, and the increasing rate is proportional to the light intensity $A$. When the voltage $V$ reaches a predefined threshold $V_{th}$, a spike will be fired and saved to the sensor register. The firing signal will wake the reset circuit to reset the voltage $V$ to zero. Controlled by the clock signal, the readout circuit reads the status of the register and resets the register. \equref{equ:if_continuous} describes how a continuous spike stream will be since the last spike fires:
\begin{align}
  \label{equ:if_continuous}
  & P(t) =
  \begin{dcases}
      & 1, \quad if \quad \int_{t^\star}^{t}CA(\tau)d\tau \ge V_{th} \\
      & 0, \quad otherwise \\
  \end{dcases}
\end{align}
where $P(t)$ is a continuous spike firing function describing whether a spike is fired at time $t$,
$V_{th}$ is predefined threshold voltage,
$C$ is the photoelectric conversion coefficient of photodiode,
$A(t)$ is light intensity at time $t$, and
$t^\star$ is the firing time of the last spike.
However, the readout spike stream is actually discrete. \equref{equ:if_concrete} formulates \equref{equ:if_continuous} into discrete form that corresponds to the output spike stream of the spike camera.

\begin{align}
  \label{equ:if_concrete}
  \begin{aligned}
      & P(n) =
      \begin{dcases}
          & 1, \quad if \quad V(n) \ge V_{th} \\
          & 0, \quad otherwise \\
      \end{dcases}
      \\
      & V(n) = V_{-}(n-1) + \int_{(n-1)\cdot \Delta T}^{n\cdot \Delta T}CA(\tau)d\tau \\
      & V_{-}(n) = V(n) - P(n)\cdot V_{th}
  \end{aligned}
\end{align}
where $P(n)$ is a discrete spike firing function describing whether a spike is fired at every integer step $n$,$\Delta T$ is the sampling time interval, $V(n)$ is the photodiode voltage (before being reset) of integrator at step $n$, and $V_{-}(n-1)$ and $V_{-}(n)$ are respectively the photodiode voltages (after being reset) of integrator at step $n-1$ and step $n$.

Due to the special output of the spiking camera, one has to apply a reconstruction method to obtain the visual image. Reported reconstruction methods for the spiking camera can mainly be divided into two categories, physical-model-based reconstruction and neural-network-model-based reconstruction.   

\textbf{Physical-model-based reconstruction:} These algorithms follow the physical integrate-and-fire rule and resolve the real light intensity from the working mechanism of the spike camera. TFI \cite{zhu2019retina} measures every interval between two adjacent spikes and takes the reciprocal of the interval as the light intensity. TFP \cite{zhu2019retina} creates a virtual exposure window with a length of $L_{win}$ to count the number of spikes $N_{spk}$ inside the window, and the $N_{spk}/L_{win}$ is considered as the light intensity. Since both algorithms are easy to implement, they can be deployed on an FPGA platform to realize real-time (20,000 FPS) reconstruction \cite{zhang2022ultra}. However, TFI produces low-quality images, especially in bright environments. TFP easily introduces motion blur for high-speed scenes, which degrades the spike camera to a conventional frame-based camera.The shortcomings of these algorithms hinder the high-quality and high-speed vision reconstruction of the spike camera.

\textbf{Neural-network-model-based reconstruction:} Since the neural networks have the ability to learn the features from the training data, they are also adopted for the image reconstruction of the spike camera. Reported Neural-network-model-based reconstruction methods can be roughly divided into Deep-Neural-Network-based (DNN-based) methods \cite{zhao2020high, zhao2021super, zhao2021spk2imgnet, zhao2022reconstructing, chen2022self, xiang2023learning, chen2023self} and Spike-Neural-Network-based (SNN-based) methods \cite{zhu2020retina, zheng2021high, zhu2023ultra, zheng2023capture}.

DNN-based methods: Since static scenes can be easily reconstructed by TFP, the reconstruction of high-speed scenes is what we should focus on. Intuitively, utilizing motion information as prior knowledge may improve reconstruction performance. Therefore, utilizing the motion information to guide the image reconstruction becomes a common characteristic of DNN-based reconstruction methods \cite{zhao2020high, zhao2021super, zhao2021spk2imgnet, zhao2022reconstructing, chen2022self, xiang2023learning, chen2023self}. \cite{zhao2020high, zhao2021spk2imgnet,zhao2022reconstructing} first extract the optical flow from the coarsely reconstructed images, and then refine the reconstruction image by the optical flow information. Differently, \cite{chen2022self, chen2023self} insert a motion inference module into a mutual learning framework to realize reconstruction, and \cite{zhao2021super, xiang2023learning} take advantage of motion estimation to achieve super-resolution reconstruction of the spike camera.

SNN-based methods:  Since the spike camera outputs spikes, intuitively, spike neuron models may be good tools to realize vision reconstruction. Therefore, \cite{zhu2020retina, zhu2023ultra} proposed a reconstruction method motivated by the bio-realistic leaky integrate-and-fire neurons and synapse connection with the spike-timing-dependent plasticity rule. Similarly, \cite{zheng2021high,zheng2023capture} adopt the short-term plasticity mechanism to recover images. 

The above-mentioned neural-network-model-based reconstruction methods can generally obtain better reconstruction performance than physical-model-based methods. However, due to computation complexity, none of them are reported for real-time vision reconstruction of the spike camera. 

In conclusion, reported physical-model-based methods are relatively simple and suitable for high-speed reconstruction, but the reconstruction quality is unsatisfactory. Reported neural-network-model-based reconstruction methods can improve the reconstruction quality, but the complexity makes them hard for real-time edge-end deployment. So, a high-quality and high-speed reconstruction algorithm is still vacant.

\section{Principle and Method}
\label{sec:method}
\subsection{Spike Stability Theorem}
Let's first back to the basic TFI and TFP methods. TFI calculates the interval of two adjacent spikes to recover light intensity. Unfortunately, two adjacent spikes contain insufficient information for accurate reconstruction because of the quantization error. TFP considers spikes during a fixed time window but easily introduces motion blur. Suppose there is a self-adaptive segmentation method that can split the spike stream into multiple short sequences corresponding to short-time static scenes, then both static and dynamic scenes can be easily reconstructed. We call those short sequences stable sequences. The stability between two adjacent stable sequences will be broken by the motion, the change of light intensity, or noise. Then the key question is how to define the stable sequence and how to determine if the stability rule is violated. To answer this question, we first introduce the spike-like stream in \defref{def:spike-like} and the stability rule of the spike-like stream in \defref{def:stability}, and then prove the spike stability theorem (\thmref{thm:stability}) that can provide a theory foundation for vision reconstruction of the spike camera.

The spike stream of the spike camera $P(n)$ consists of 1 (firing value) and 0 (resting value). To introduce our spike stability theorem, we extend the firing value to be any number $e_1$ and the resting value to be any value $e_2$. A digital sequence consisting of resting value and firing value is called a spike-like stream. The definition of the spike-like stream is given in \defref{def:spike-like}.

\begin{definition}
  \label{def:spike-like}
  Given an integrator with threshold $M^{th}$, check the status of integrator $M$ at every integer step n. If $M(n) >= M^{th}$, appends firing value $e_1$ to stream $S$, otherwise appends resting value $e_2$ to stream $S$. Such stream $S$ is a spike-like stream. Specifically, if $e_1=1$ and $e_2=0$, $S$ is a spike stream.
\end{definition}

The mathematical form of the spike-like stream can be described as \equref{equ:if_spike-like}.
\begin{align}
  \label{equ:if_spike-like}
  \begin{aligned}
      & S(n) =
      \begin{dcases}
          & e_1, \quad if \quad M(n) \ge M^{th} \\
          & e_2, \quad otherwise \\
      \end{dcases}
      \\
      & M(n) = M_{-}(n-1) + Q(n)\\
      & M_{-}(n) = M(n) - \frac{S\left( n \right)-{{e}_{2}}}{{{e}_{1}}-{{e}_{2}}}{{M}^{th}}
  \end{aligned}
\end{align}
where $M(n)$ is the status (before being reset) of the integrator at step $n$, $e_1$ is the firing value, $e_2$ is the resting value, $M^{th}$ is the comparison threshold of integrator, $M_{-}(n-1)$ and $M_{-}(n)$ are respectively the status (after being reset) of integrator at step $n-1$ and step $n$, and $Q(n)$ is the accumulation rate of integrator. Here we assume the accumulation rate $Q(n)$ does not change between steps $n$ and $n-1$.

\begin{definition}
    \label{def:stability}
    Given a spike-like stream $S_0$, if $S_0$ consists of one element (firing value $e_1$), or two elements (firing value $e_1$ and resting value $e_2$) with $|e_2-e_1|=1$, $S_0$ is considered as a zero-order stable stream. We can calculate the interval between each adjacent $e_1$ and generate its interval stream $S_1$. Then we have,
    \begin{small}
        \begin{align}
            \label{equ:spike-stability-1}
            &
            \begin{aligned}
                & S_0 = \{S_0(1), S_0(2),..., S_0(m), ...\} \\
                & S_1 = \{T_0(1), T_0(2),..., T_0(m), ...\}
            \end{aligned}
        \end{align}
    \end{small}
    where $S_0(m)$ represents the $m_{th}$ element of $S_0$ (it is $e_1$ or $e_2$), $S_1$ is the interval stream of $S_0$, and $T_0(m)$ is the interval between $m^{th}$ and $(m+1)^{th}$ firing values of $S_0$. 
    
    If $S_1$ also meets the zero-order stable rule (in other words, see an arbitrary element of $S_1$ as a new firing value $e_1$, the other elements must be $e_1$ or $e_2$ with $|e_2-e_1|=1$), then we say $S_0$ is a first-order stable stream. We can continue the above operations $n$ times until $S_n$ violates the zero-order stability rule. That is, we have,
    \begin{small}
        \begin{align}
            \label{equ:spike-stability-2}
            &
            \begin{aligned}
                & S_0 = \{ S_0(1), S_0(2),..., S_0(m), ...\} \\
                & S_1 = \{ T_0(1), T_0(2),..., T_0(m), ...\} \\
                & ... \\
                & S_n = \{ T_{n-1}(1), T_{n-1}(2),..., T_{n-1}(m), ...\}
            \end{aligned}
        \end{align}
    \end{small}
    where $T_{n-1}(m)$ is the interval between $m^{th}$ and $(m+1)^{th}$ firing values of $S_{n-1}$. Then, we say $S_0$ is $n$-order stable stream. If $n$ approaches infinity or $S_n$ is empty, then we say $S_0$ is an absolutely stable stream.
\end{definition}

Stability is a key concept in this paper since we found an interesting conclusion between the accumulation rate $Q$ and stability described in \thmref{thm:stability}.

\begin{theorem}
    \label{thm:stability}
    Given a fixed accumulation rate $Q_0$, spike-like stream $S_0$ is an absolutely stable stream.
\end{theorem}

Since the spike stream is a special case of the spike-like stream, \thmref{thm:stability} can be applied to the spike camera. The complete proof of \thmref{thm:stability} see the Appendix. We use the conclusion directly in the rest of this paper. The conclusion is simple but powerful because it splits the reconstruction into two stages: 1. segment spike stream into multiple stable sequences; 2. calculate light intensity within the stable sequence. As discussed in \secref{sec:relate}, for a static scene (implying fixed accumulation rate $Q_0$), reconstruction is a relatively easy job. So we focus on how to segment the spike stream. Note that, \thmref{thm:stability} proposes a sufficient condition. Using \thmref{thm:stability} for image reconstruction assumes that the change rate of light intensity is not too fast. Strictly, the change rate of light intensity should be slower than the sampling rate of the spike camera. Actually, if the light intensity changes faster than the sampling rate, the output of the spike camera becomes unbelievable, making the reconstruction an ill-posed problem. In fact, all reconstruction methods referring to \secref{sec:intro} implicitly follow the assumption. \cite{zhu2023ultra} mentions this challenging issue and provides several solutions for such conditions.

\begin{figure*}[htbp]
  \centering
  \includegraphics[width=\textwidth]{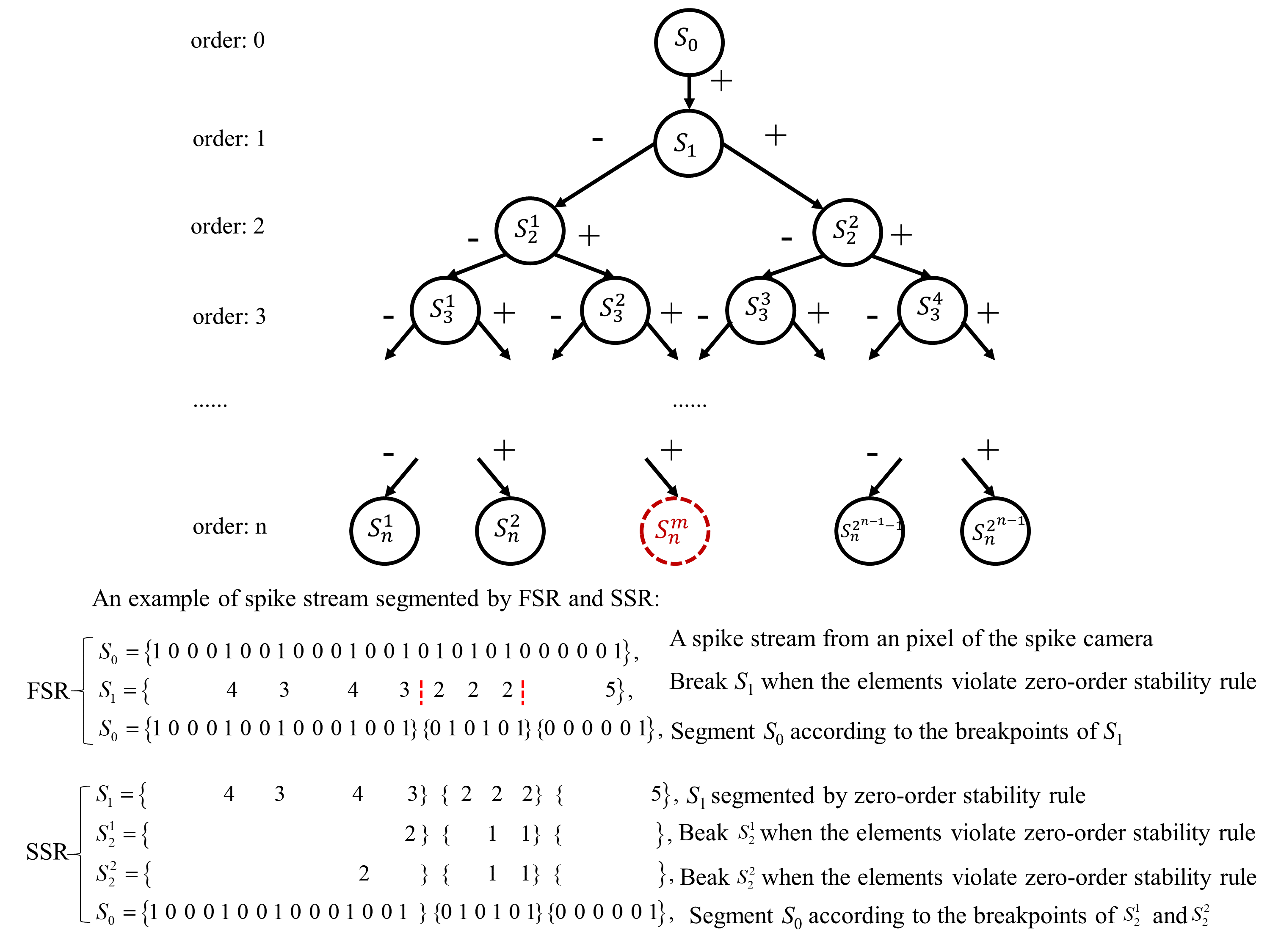}
  \caption{The interval sequences needed to calculate when segmenting a spike stream $S_0$ by n-order stable judgment. Red dotted circle indicates that the interval sequence $S_n^m$ is empty. The '-' indicates selecting the smaller value as the firing value, and the '+' indicates selecting the larger value as the firing value. An example of spike stream $S_0$ segmented by FSR (First-order Stability Reconstruction) and SSR (Second-order Stability Reconstruction).}
  \label{fig:path}
\end{figure*}

\subsection{Reconstruction Algorithms}

According to the above analysis, the vision reconstruction of the spike camera can be realized by two steps: spike stream segmentation and light intensity calculation. Next, we will introduce the two steps in detail.  

\textbf{Spike stream segmentation:} Since a static scene implies a fixed accumulation rate, the key step of reconstruction is to segment the spike stream of the spike camera into multiple absolutely stable streams. We suppose $S_0$ is a spike stream from a pixel of the spike camera. Theoretically, to determine whether $S_0$ is an absolutely stable stream, one has to calculate interval sequences $S_1$, $S_2^1$ and $S_2^2$,..., as shown in \figref{fig:path}, and judge them whether violate the zero-order stability rule. The calculation complexity to complete the judgment process will be $O(2^{n-1})$ where $n$ is the order that makes $S_n^m$ empty. However, we find, for most cases, the segmentation result of n-order is the same as that of 1-order. So, in practice, restricting $n$ to 2 is enough to obtain acceptable reconstruction results. When $n=1$, we only need to calculate $S_1$ and judge whether it violates the zero-order stability rule. We call the corresponding method first-order stability reconstruction (FSR). When $n=2$, we call the corresponding method second-order stability reconstruction (SSR). The calculation complexity of FSR and SSR is low, which benefits the edge-end deployment. To make the segmentation process clearer, here we give a spike segmentation example of FSR and SSR in \figref{fig:path}. For FSR, we first obtain $S_1$ from $S_0$ and then break the $S_1$ when its elements violate the zero-order stability rule (judge if a sequence only consists of one element or two elements with their difference equal to 1). According to the breakpoints of $S_1$, we can easily segment $S_0$ into three parts. For SSR, we further calculate the $S_2^{1}$ and $S_2^{2}$ based on the segmented $S_1$ from FSR, and then break them when the elements violate the zero-order stability rule. For the example in \figref{fig:path}, as all the elements of $S_2^{1}$ and $S_2^{2}$ do not violate the zero-order stability rule, no additional breakpoints are added in $S_2^{1}$ and $S_2^{2}$. So the $S_0$ segmented by SSR is the same as that of FSR. One can see that the segmentation is completed by mainly judging the interval sequences $S_1$, $S_2^{1}$, and $S_2^{2}$ whether violate the zero-order stability rule, there is no need to set any parameters to complete the segmentation.

\textbf{Light intensity calculation:} After the segmentation, we can calculate the light intensity using the segmented stable spike streams. For a stable spike stream, the actual interval of spikes is the mean of all readout intervals. For that reason, the mean spike interval is used to approach the actual interval. Given the approached interval, we recover the light intensity similar to TFI \cite{zhu2019retina}. The final pixel intensity is the reciprocal of the approached interval multiplied by 255 (8-bit image). The intensity reconstruction of a segmented stable spike stream $S_0$ can be referred to \equref{equ:approx_isi}:
\begin{equation}
    \label{equ:approx_isi}
    I=\frac{255}{\frac{1}{N}\sum_i{T_0(i)}}
\end{equation}
where $I$ is the reconstructed pixel intensity, $T_0(i)$ is the interval value between $i^{th}$ spike and $(i+1)^{th}$ spike in segmented $S_0$, and $N$ is the total number of the interval values.

\begin{figure}[t!]
  \includegraphics[width=\columnwidth]{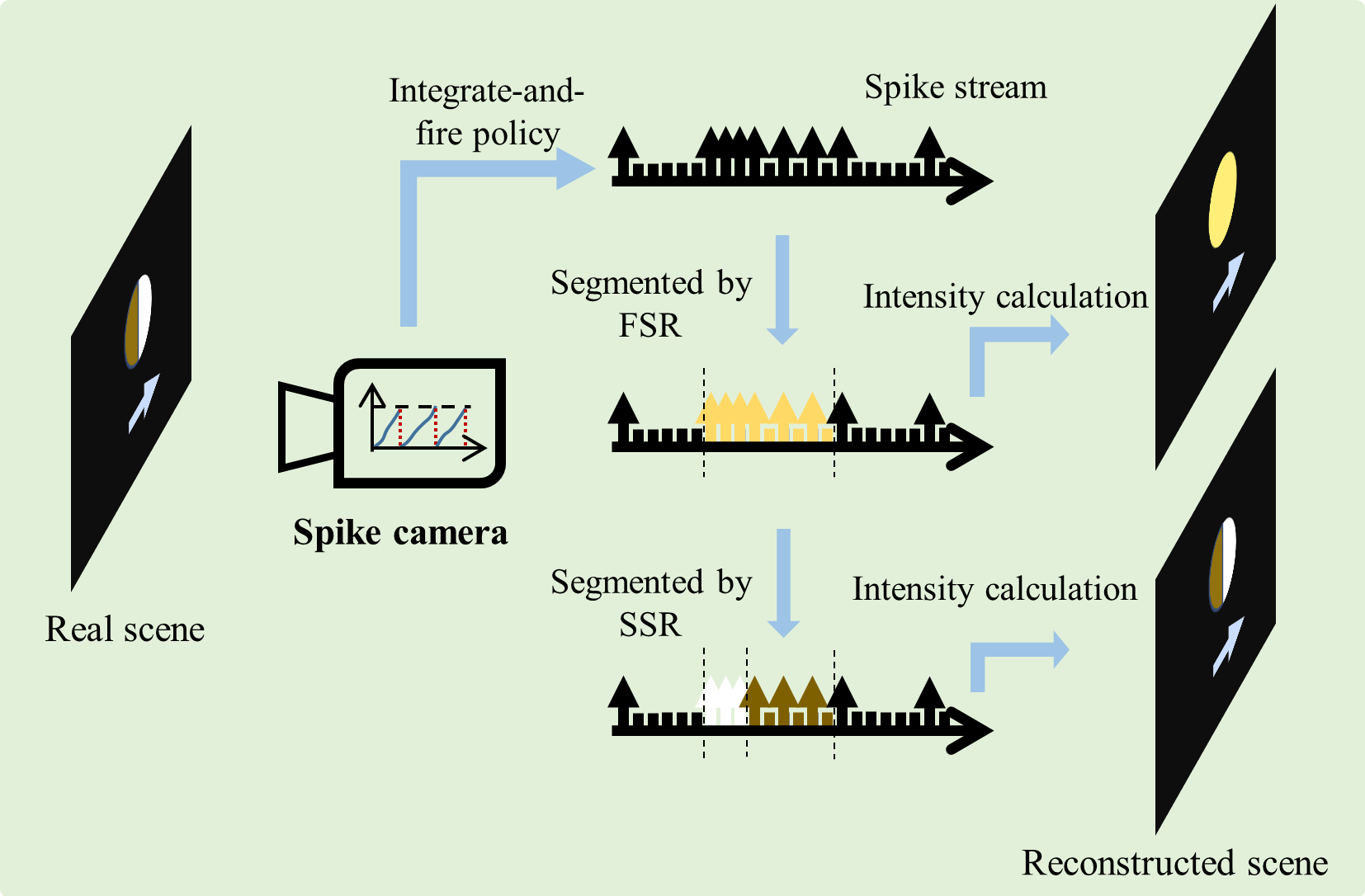}
  \caption{Visual representation of the simplified workflow of FSR (First-order Stability Reconstruction) and SSR (Second-order Stability Reconstruction).}
  \label{fig:simplified_workflow}
\end{figure}

To make the core idea of the algorithms more intuitive to readers, \figref{fig:simplified_workflow} gives a visual representation of the simplified workflow of FSR and SSR. Both FSR and SSR include two steps, spike stream segmentation and light intensity calculation. The difference between FSR and SSR is during the spike stream segmentation. SSR may break a spike stream segmented by FSR into shorter spike streams. By the way, it is worth pointing out that there are no parameters to set for the FSR and SSR, which is an outstanding advantage compared with reported algorithms.

In practice, due to the influence of noise, the stable sequence $S_0$ segmented by FSR and SSR may be different. Normally, $S_0$ segmented by SSR will have a finer reconstruction texture as it guarantees higher-order stability than FSR. Of course, SSR will cost more calculation resources as it needs second-order stability determination. We will discuss the performance difference between FSR and SSR in \secref{sec: Experimental results}.

\subsection{FPGA Implementation}
\begin{figure*}[htbp]
  \centering
  \includegraphics[width=\textwidth]{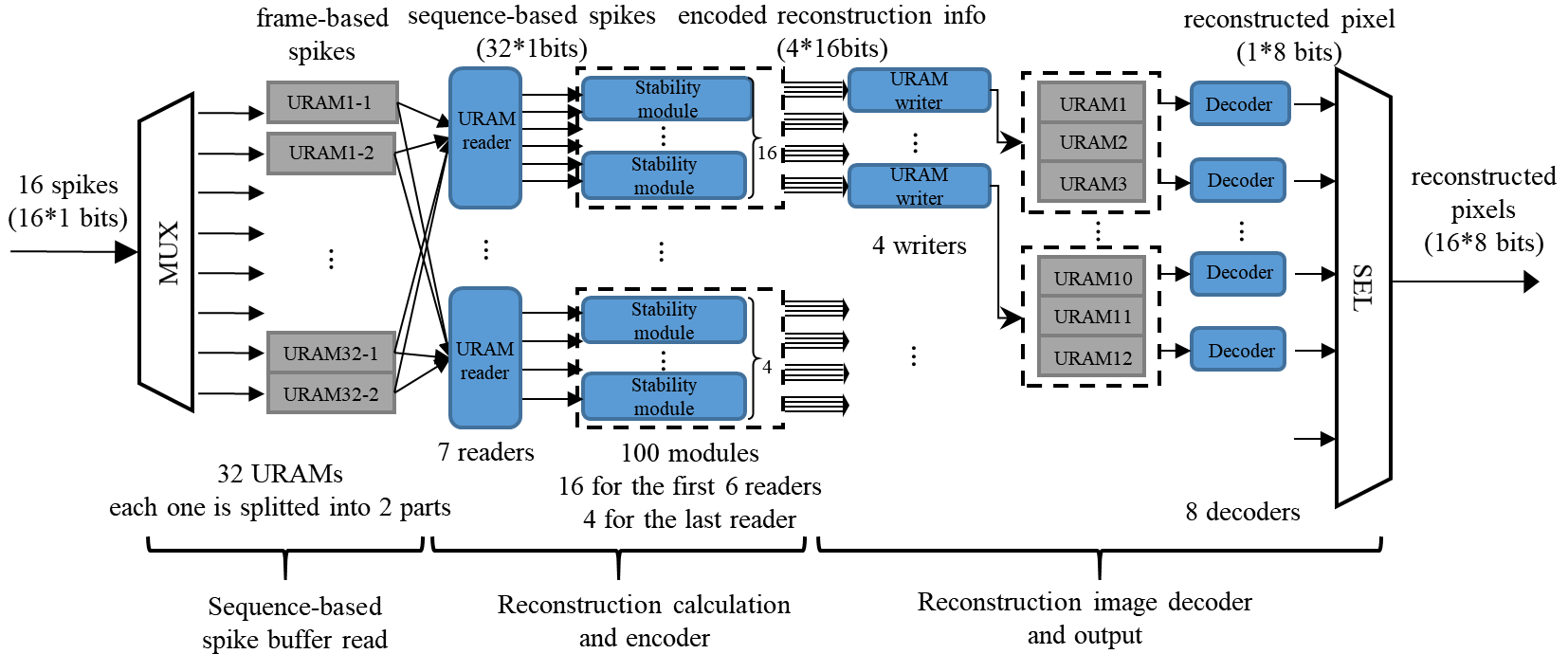}
  \caption{Overview of FPGA-implemented reconstruction.}
  \label{fig:fpga_overview}
\end{figure*}

The FPGA-implemented reconstruction module can be divided into three parts: sequence-based spike buffer read, reconstruction calculation and encoder, and reconstruction image decoder and output. See \figref{fig:fpga_overview}.

\textbf{Overview:} The Xilinx VCU118 evaluation board runs at a clock frequency of 150 MHz. Since FSR and SSR are sequence-based reconstruction methods, we could not reconstruct the image from incoming spikes directly. The incoming spikes (from 16 adjacent pixels in one spike frame) are first cached in a sequential form at a specific URAM. The reconstruction process starts when 32-frame spikes are entirely cached in URAM. The reconstruction process will be done in less than $32 \times 50\mu s=1.6ms$ to avoid data corruption. It needs to be pointed out that the $ 50\mu s$ is related to 20,000 spike frames per second generated by the spike camera. In other words, the processing speed of FPGA-implemented reconstruction is 20,000 FPS.

\textbf{Sequence-based spike buffer read:} Triggered by the rising edge of the clock, the multiplexer receives 16 spikes from the spike camera (1 bit per pixel) and allocates them to specific URAM. These pixels are transferred row by row until the whole spike frame is covered in frame-based format. Because FSR/SSR algorithms rely on sequence-based spikes, the incoming spikes are first cached in 32 URAMs. To make sure the write and read of URAMs can be simultaneously executed, these 32 URAMs are partitioned into 64 blocks. When the writers operate on the first 32 blocks, the readers will read the last 32 blocks and vice versa.

\textbf{Reconstruction calculation and encoder:} There are 7 URAM readers to read the cached spikes, and the spikes will be sent to the stability modules in the form of 32 time-continuous spikes from one pixel. One hundred stability modules are instantiated to process all the 10,000 ($400\times250$) pixels of the spike camera. Since we have 7 URAM readers, each URAM reader is responsible for 16 stability modules except the last URAM reader for the remaining 4 stability modules. The output of stability modules is the encoded reconstruction information. Specifically, encoded reconstruction information (16-bit) is composed of two 8-bit variables: the number of stability duration frames and the reconstructed pixel intensity. Since reconstruction values are invariant within a stable sequence, this encoded method largely reduces the usage of RAM.

\begin{figure*}[htbp]
    \centering
    \includegraphics[width=0.9\textwidth]{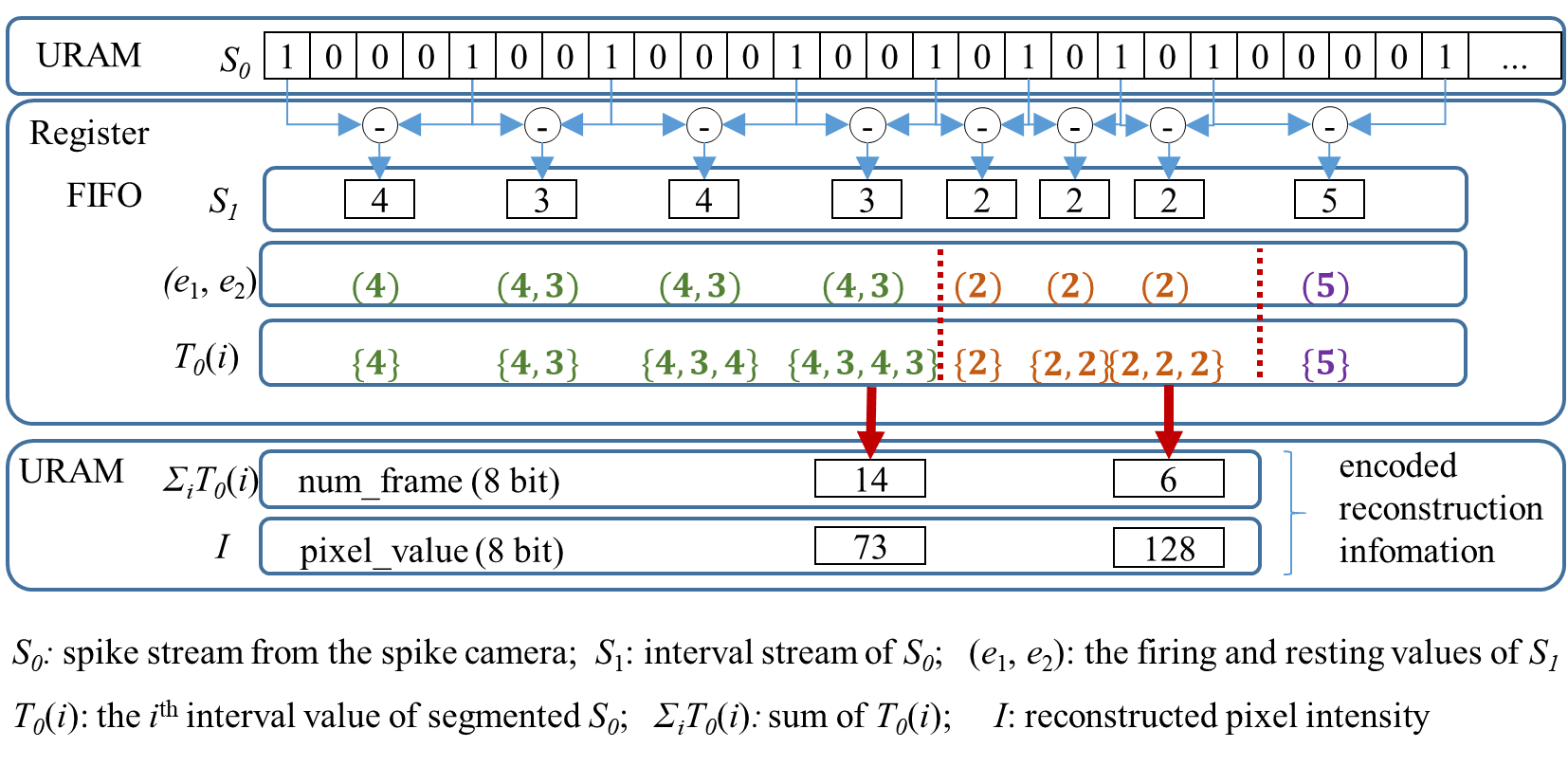}
    \caption{The detail of FPGA-implemented stability module (FSR).}
    \label{fig:fpga_fsr}
\end{figure*}

More details of the stability module are presented in \figref{fig:fpga_fsr}. To make the calculation process of the stability module clear, we still take the example of the above-mentioned FSR as an example: the URAM readers read the spike stream $S_0$ from the spike camera, and then intervals stream $S_1$ is calculated and sent to a FIFO. Next, the processor will temporarily store the firing and resting values $(e_1,e_2)$. When a new element coming from $S_1$ makes  $(e_1,e_2)$ satisfy the zero-order stability rule, the element will be stored as the $i^{th}$ interval value of the segmented $S_0$, $T_0(i)$. When a new element coming from $S_1$ makes $(e_1,e_2)$ violate the zero-order stability rule, both $\sum_i{T_0(i)}$(equals to the number of stability duration frames) and $I$(the reconstructed pixel intensity) will be calculated and popped out to generate the encoded reconstruction information, and both the $(e_1,e_2)$ and $T_0(i)$ will be reset with the new element at the same time.

\textbf{Reconstruction image decoder and output:} This part is used to decode the encoded information and output reconstruction results. Before the decoder, the encoded information needs to be cashed into URAMs. Four URAM writers are designed to pack and transfer the encoded information from four adjacent pixels to URAMs simultaneously. To accommodate a long stable sequence, three URAMs are used for each writer. Since the encoded information does not match the output protocol. Eight decoders are designed to align the reconstruction information to the specific output format. The final output pixel values are represented by 8 bits.

\section{Experimental Result}

\subsection{Datasets}
To compare the performances of different algorithms, we test the algorithms on two different datasets, including PKU-Spike-High-Speed \cite{zhu2019retina} and our SpikeCityPCL. PKU-Spike-High-Speed focuses on high-speed moving objects (up to 350 km/h) whose textures and structures are relatively simple. To better measure the reconstruction quality of different algorithms, we provide a new SpikeCityPCL as compensation (The dataset can be found at \url{https://openi.pcl.ac.cn/BDIP/SpikeCityPCL}). This new dataset was collected by a portable spike camera with a sampling rate of 20,000Hz and a spatial resolution of ($400\times250$). \textbf{SpikeCityPCL:} SpikeCityPCL records scenes with rich texture information, and there are 248 scenes in total. Visualization examples can be seen in \figref{fig:experiments} and \figref{fig:comparison}.

\subsection{Experimental results}
\label{sec: Experimental results}
To evaluate the performances of our methods, we compare them with six representative vision reconstruction methods, including Physical-model-based methods (TFI(texture from the inter-spike interval) \cite{zhu2019retina}, TFP (texture from playback) \cite{zhu2019retina} ) and Neural-network-model-based methods (Spk2Img (spike-to-image neural network) \cite{zhao2021spk2imgnet}, SSML(Self-Supervised Mutual Learning) \cite{chen2022self}, SNM (spiking neuron model) \cite{zhu2023ultra}, TFSTP(texture from short-term plasticity)\cite{zheng2023capture}). The Spk2img and SSML are DNN-based methods, and the SNM and TFSTP are SNN-based methods. The parameters of Spk2Img, SSML, SNM, and TFSTP are set according to the default parameters given in the original papers. As the window length $L_{win}$ of TFP has impacts on the reconstruction, we use two kinds of $L_{win}$s (TFP-32 ($L_{win}=32$) and TFP-64  ($L_{win}=64$)) that own better reconstruction performance for comparison. We evaluate the performances of these algorithms from two perspectives: reconstruction quality and reconstruction speed.
 
\textbf{Reconstruction quality:} Three commonly used no-reference (NR) image quality assessment metrics, blind/referenceless image spatial quality evaluator(BRISQUE) \cite{Mittal2012noRefernce}, perception index(PI) \cite{blau20182018}, and two-dimensional entropy (TE) \cite{xi1999autofocusing}, are used to evaluate the performances of reconstruction methods. BRISQUE quantifies possible losses of naturalness in the image due to the presence of distortions by using scene statistics of locally normalized luminance coefficients, and a smaller BRISQUE indicates higher image quality. PI evaluates the perceptual quality of the image by combining NIQE (an index that estimates image quality by measuring the deviations from statistical regularities of natural images) and Ma (an index that quantifies the image quality from three statistical properties, including local/global frequency variations, and spatial discontinuity), and a smaller PI indicates better perceptual quality. TE measures the amount of information in the image using both the gray value of a pixel and its local average gray value, and a larger TE means more information.

We calculate the average values of BRISQUE, PI, and TE on the two datasets, as shown in \tabref{tab:qualitative_metrics}. From the BRISQUE in \tabref{tab:qualitative_metrics}, one can see that the SNM gets the best results, and our SSR ranks second on both datasets. From the perspective of PI, the SSML has the best performance, and our methods rank in the middle. As for TE, our methods and SSML have the highest scores in the PKU-Spike-High-Speed dataset. In the SpikeCityPCL dataset, the SNM has the best TE, and our SSR ranks third among all the methods. It is worth pointing out that our methods (FSR and SSR) obtain better BRISQUE, PI, and TE than all the physical-model-based methods (TFI, TFP-32, and TFP-64) and the Neural-network-model-based TFSTP. As different metrics have different emphases on image quality evaluation, no one method can outperform other methods on all the metrics. Overall, our methods outperform TFI, TFP-32, TFP-64, and TFSTP, and are close to the levels of the state-of-the-art methods(Spk2Img, SSML, and SNM).

\begin{table*}[!t]
\caption{\\Quantitative reconstruction quality comparison of different methods on PKU-Spike-High-Speed and SpikeCityPCL datasets\label{tab:qualitative_metrics}}
\centering
\begin{tabular*}{\textwidth}{@{\extracolsep{\fill}} c|ccc|ccc}
\hline\hline
\multicolumn{1}{c|}{\multirow{2}{*}{\textbf{Methods}}} & \multicolumn{3}{c|}{\textbf{PKU-Spike-High-Speed}} & \multicolumn{3}{c}{\textbf{SpikeCityPCL}}\\
\cline{2-7}
& \textbf{BRISQUE}($\downarrow$) & \textbf{PI}($\downarrow$) & \textbf{TE}($\uparrow$) & \textbf{BRISQUE}($\downarrow$) & \textbf{PI}($\downarrow$) & \textbf{TE}($\uparrow$)  \\
\hline
TFI \cite{zhu2019retina}           & 32.2 & 6.3 & 7.1 & 31.5 &6.3 &6.5 \\
TFP-32 \cite{zhu2019retina}        & 40.0 & 6.7 & 7.0 & 58.6 &10.7 &6.2 \\
TFP-64 \cite{zhu2019retina}        & 35.3 & 6.4 & 7.4 & 39.4 &8.2 &6.4\\
Spk2Img \cite{zhao2021spk2imgnet}  & 29.2 & 5.0 & 8.6 & 40.1 & 5.1 &7.3 \\
SSML \cite{chen2022self}           & 30.9 & \textbf{4.1} & \textbf{9.2} & 37.0 &\textbf{4.3} & 8.4\\
SNM \cite{zhu2023ultra}            & \textbf{15.6} & 4.5 & 9.1 & \textbf{17.3} &4.6 &\textbf{9.1}  \\
TFSTP \cite{zheng2023capture}      & 46.7 & 6.7 & 7.5 & 45.8 &6.9 & 7.0  \\
FSR(ours)                          & 20.5 &6.1  & \textbf{9.2} &27.0 &6.1 & 7.9\\
SSR(ours)                          & 19.3 & 6.0 & \textbf{9.2} &24.5 &6.4 & 8.1 \\
\hline\hline
\end{tabular*}
\end{table*}

\begin{figure*}[htbp]
    \begin{tabularx}{\textwidth}{cccccc}
         & rotation & car & train & road & building \\
         \rotatebox{90}{\quad Spike} &
         \includegraphics[width=0.15\textwidth]{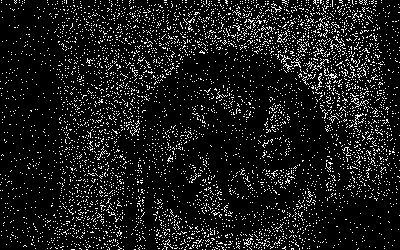} & \includegraphics[width=0.15\textwidth]{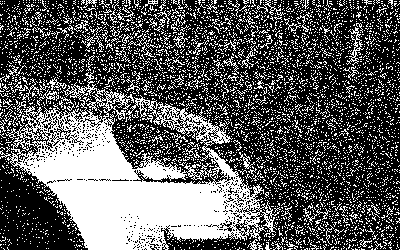} & \includegraphics[width=0.15\textwidth]{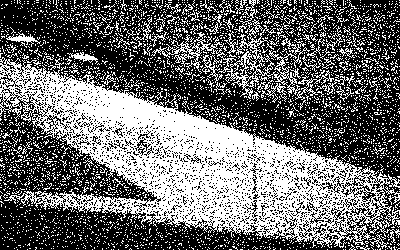} & \includegraphics[width=0.15\textwidth]{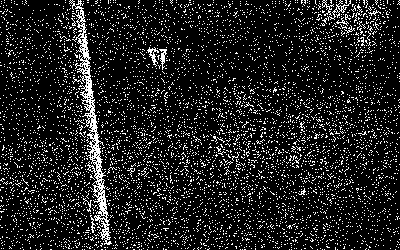} & \includegraphics[width=0.15\textwidth]{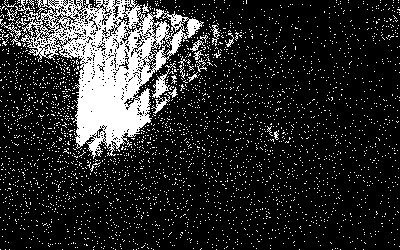} \\
         & (BRISQUE, PI, TE) & (BRISQUE, PI, TE) & (BRISQUE, PI, TE) & (BRISQUE, PI, TE) & (BRISQUE, PI, TE)\\
         \rotatebox{90}{\quad \quad TFI} &
         \includegraphics[width=0.15\textwidth]{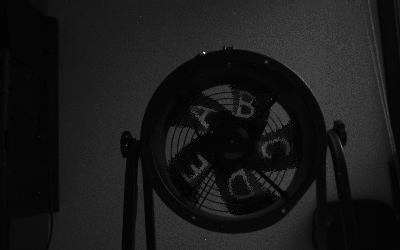} & \includegraphics[width=0.15\textwidth]{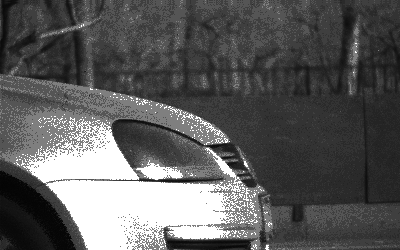} & \includegraphics[width=0.15\textwidth]{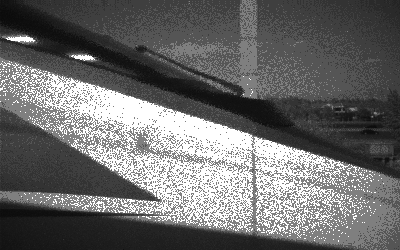} & \includegraphics[width=0.15\textwidth]{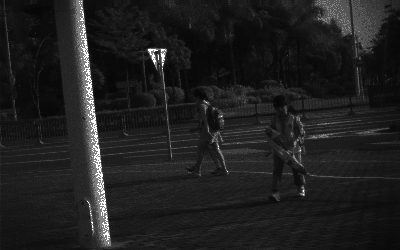} & \includegraphics[width=0.15\textwidth]{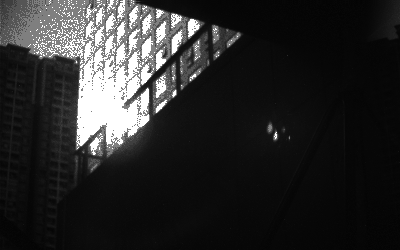} \\
         & (11.0, 7.2, 7.4) & (53.2, 5.7, 7.1) & (69.3, 6.8, 7.2) & (24.0, 3.3, 7.9) &(16.9, 4.3, 6.7)\\
         \rotatebox{90}{\quad TFP-32} &
         \includegraphics[width=0.15\textwidth]{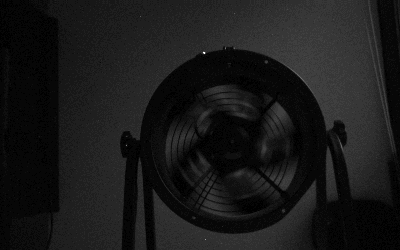} & \includegraphics[width=0.15\textwidth]{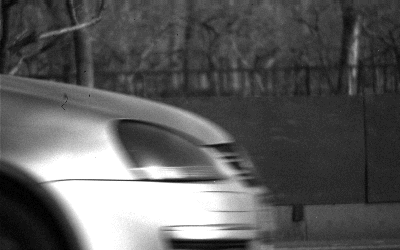} & \includegraphics[width=0.15\textwidth]{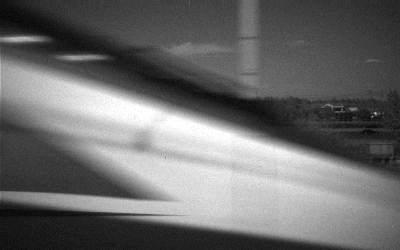} & \includegraphics[width=0.15\textwidth]{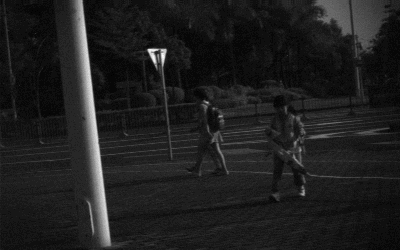} & \includegraphics[width=0.15\textwidth]{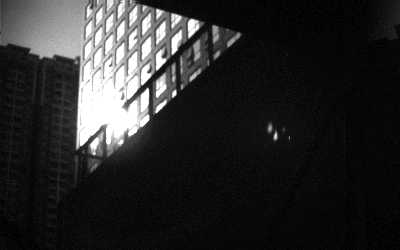} \\
         & (43.7, 7.7, 6.5) & (26.5, 6.3, 8.0) & (14.9, 8.2, 8.7) & (38.7, 5.7, 6.9) &(57.2, 5.7, 6.6)\\
         \rotatebox{90}{\quad TFP-64} &
         \includegraphics[width=0.15\textwidth]{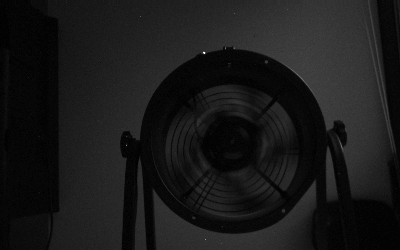} & \includegraphics[width=0.15\textwidth]{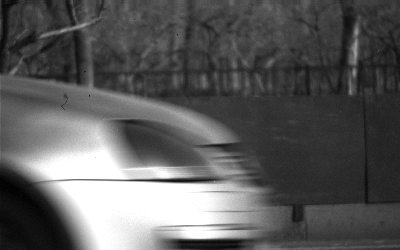} & \includegraphics[width=0.15\textwidth]{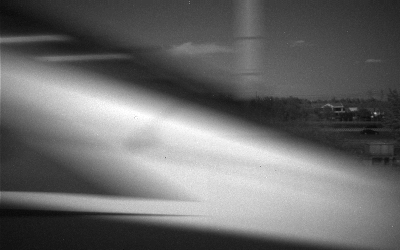} & \includegraphics[width=0.15\textwidth]{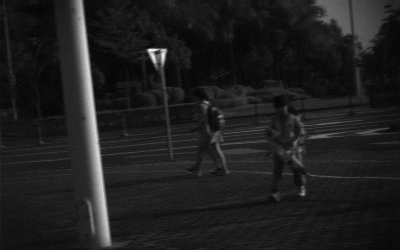} & \includegraphics[width=0.15\textwidth]{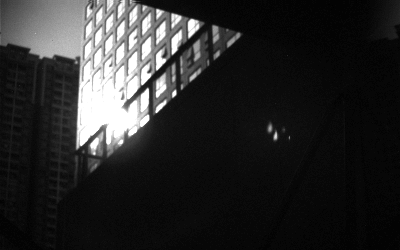} \\
         & (31.0, 6.9, 6.8) & (25.8, 5.8, 8.6) & (5.9, 9.1, 9.2) & (31.8, 6.0, 7.3) &(43.2, 5.2, 6.9)\\
         \rotatebox{90}{\quad Spk2Img} &
         \includegraphics[width=0.15\textwidth]{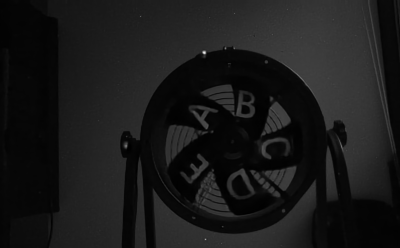} & \includegraphics[width=0.15\textwidth]{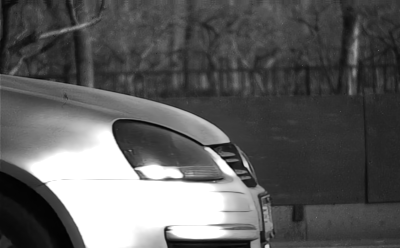} & \includegraphics[width=0.15\textwidth]{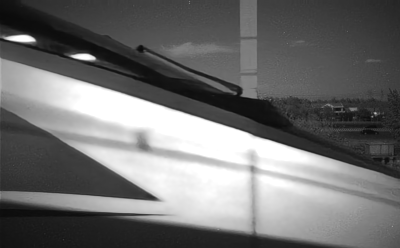} & \includegraphics[width=0.15\textwidth]{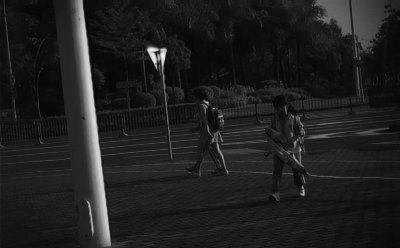} & \includegraphics[width=0.15\textwidth]{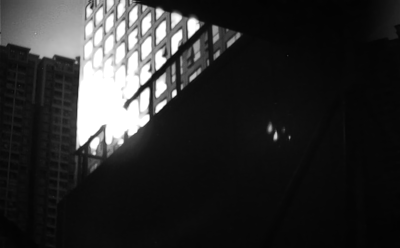} \\
          & (24.4, 5.4, 7.6) & (29.0, 3.1, 9.6) & (22.8, 5.4, 10.2) & (35.4, 3.7, 9.4) &(49.9, 5.0, 7.1)\\
         \rotatebox{90}{\quad SSML} &
         \includegraphics[width=0.15\textwidth]{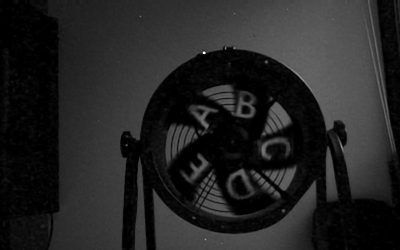} & \includegraphics[width=0.15\textwidth]{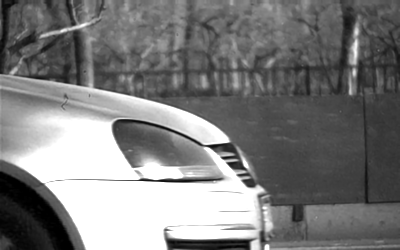} & \includegraphics[width=0.15\textwidth]{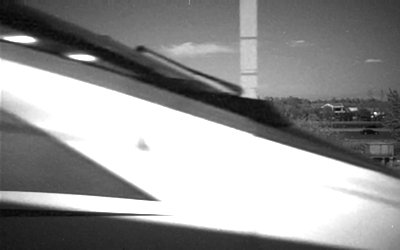} & \includegraphics[width=0.15\textwidth]{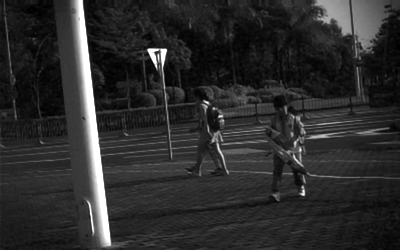} & \includegraphics[width=0.15\textwidth]{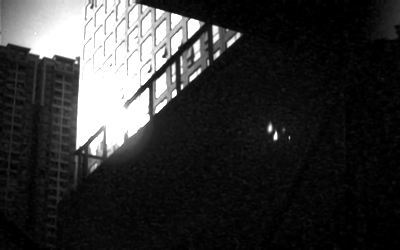} \\
         & (25.0, 3.3, 8.9) & (40.9, 4.0, 9.9) & (36.1, 6.5, 8.7) & (29.2, 3.0, 10.3) &(39.3, 3.2, 8.5)\\
         \rotatebox{90}{\quad SNM} &
         \includegraphics[width=0.15\textwidth]{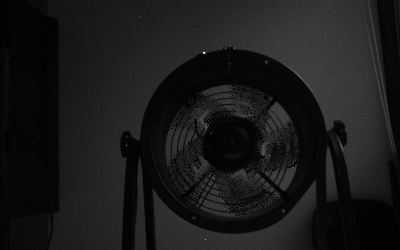} & \includegraphics[width=0.15\textwidth]{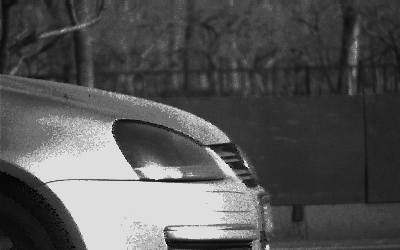} & \includegraphics[width=0.15\textwidth]{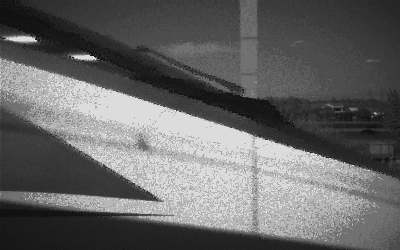} & \includegraphics[width=0.15\textwidth]{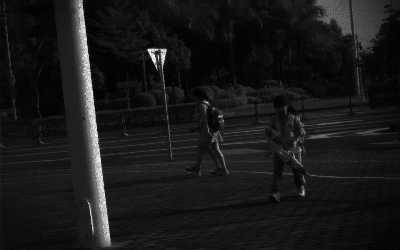} & \includegraphics[width=0.15\textwidth]{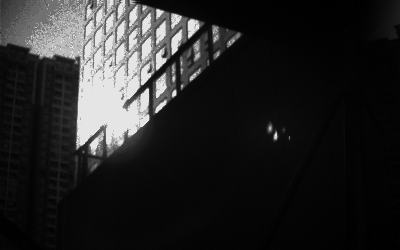} \\
         & (1.6, 7.0, 8.5) & (19.3, 3.0, 9.1) & (12.3, 3.3, 10.2) & (28.1, 3.7, 9.0) &(29.2, 3.6, 7.2)\\
         \rotatebox{90}{\quad TFSTP} &
         \includegraphics[width=0.15\textwidth]{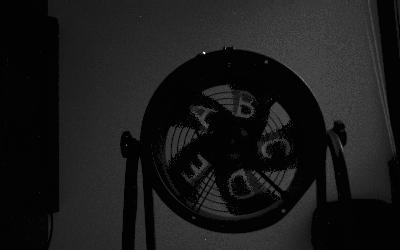} & \includegraphics[width=0.15\textwidth]{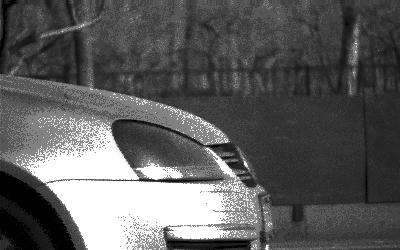} & \includegraphics[width=0.15\textwidth]{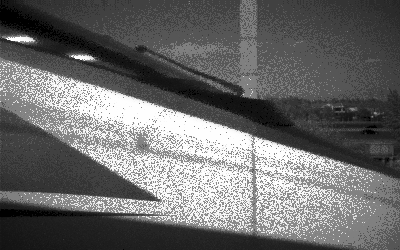} & \includegraphics[width=0.15\textwidth]{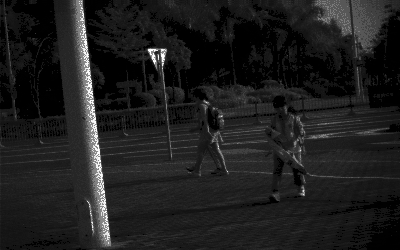} & \includegraphics[width=0.15\textwidth]{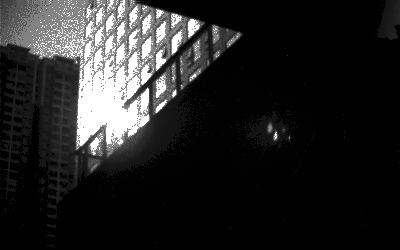} \\
         & (73.9, 9.3, 7.9) & (48.5, 5.5, 8.6) & (64.6, 6.5, 10.0) & (41.9, 4.2, 8.1) &(78.1, 6.0, 5.5)\\
         \rotatebox{90}{\quad FSR(ours)} &
         \includegraphics[width=0.15\textwidth]{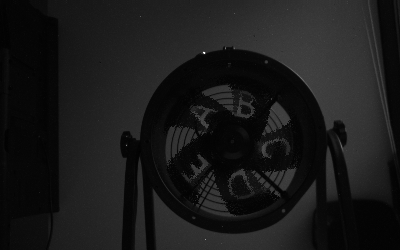} & \includegraphics[width=0.15\textwidth]{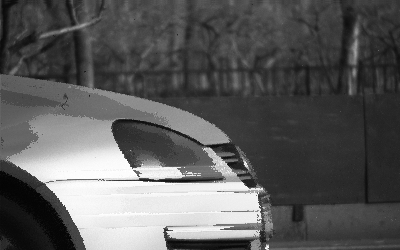} & \includegraphics[width=0.15\textwidth]{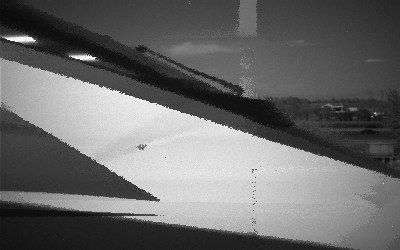} & \includegraphics[width=0.15\textwidth]{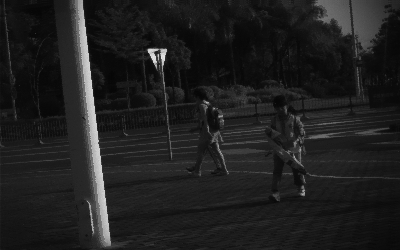} & \includegraphics[width=0.15\textwidth]{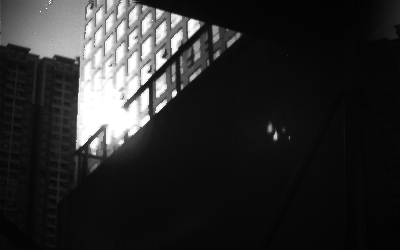} \\
         & (6.7, 7.4, 8.3) & (25.0, 3.7, 10.7) & (11.7, 6.5, 11.4) & (27.9, 3.4, 9.2) &(30.1, 4.5, 7.9)\\
         \rotatebox{90}{\quad SSR(ours)} &
         \includegraphics[width=0.15\textwidth]{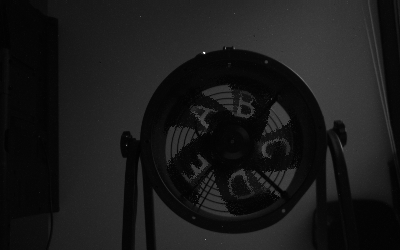} & \includegraphics[width=0.15\textwidth]{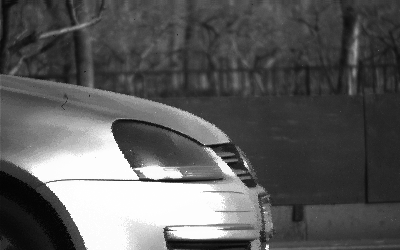} & \includegraphics[width=0.15\textwidth]{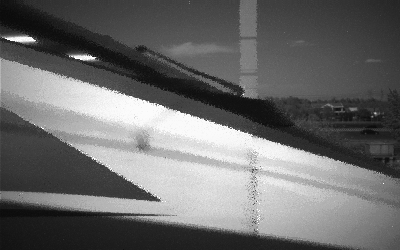} & \includegraphics[width=0.15\textwidth]{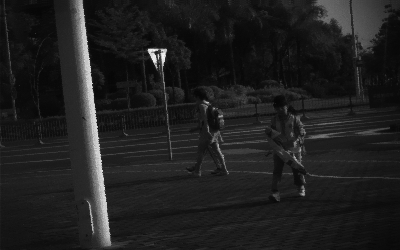} & \includegraphics[width=0.15\textwidth]{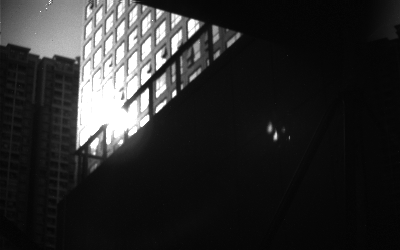}\\
         & (6.7, 7.4, 8.3) & (25.9, 3.7, 10.3) & (8.5, 6.3, 11.5) & (27.9, 3.6, 9.3) &(25.9, 4.2, 7.8)
    \end{tabularx}
    \caption{Images reconstructed by different algorithms under some typical high-speed moving and rich-texture scenes. (rotation, car, and train are from dataset PKU-Spike-High-Speed, and road and building are from dataset SpikeCityPCL)}
    \label{fig:experiments}
\end{figure*}

To make the reconstruction results more intuitive, some typical high-speed moving and rich-texture scenes are displayed in \figref{fig:experiments}. From \figref{fig:experiments}, one can find that the TFI can well reconstruct the outlines of the high-speed scenes ('rotation', 'car', and 'train'), but it has severe noise in all the scenes. Compared with TFI, TFP-32 and TFP-64 greatly reduce the noise, but have obvious motion blur. The larger the $L_{win}$ of TFP, the worse the motion blur. Almost all the Neural-network-model-based methods (Spk2Img, SSML, SNM, and TFSTP) can avoid motion blur, but there are still some scenes where they can not obtain stable results. For example, the Spk2Img outputs some defects on the train (see the 'train' scene). The SSML obtains an overexposure at the windows in the 'building' scene. The SNM can not reconstruct clearly the letters(A, B, C, D, E) in the 'rotation' scene. The TFSTP does not solve the problem of random noise (see 'car' and 'train' scenes). In contrast, our methods (FSR and SSR) realize relatively stable reconstruction without obvious blur and noise in all the scenes.  

\begin{figure}[!t]
    \small
    \begin{tabularx}{\columnwidth}{m{0.001\columnwidth}m{0.33\columnwidth}m{0.33\columnwidth}m{0.05\columnwidth}}
         & \multicolumn{1}{c}{FSR} & \multicolumn{1}{c}{SSR} & \\
         \rotatebox{90}{\quad grocery} &
         \includegraphics[height=0.23\columnwidth]{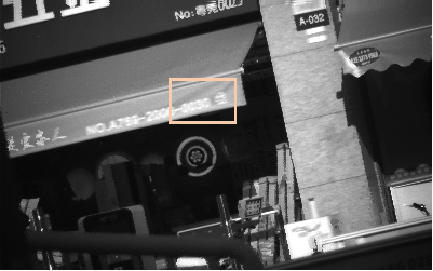} & \includegraphics[height=0.23\columnwidth]{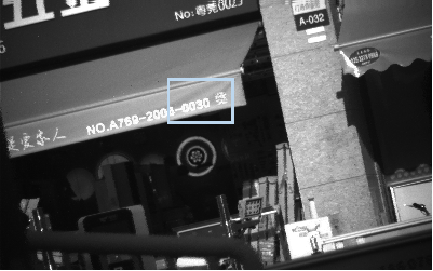} & \includegraphics[height=0.23\columnwidth]{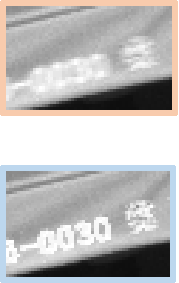} \\
         & \multicolumn{1}{c}{(BRISQUE, PI, TE)} & \multicolumn{1}{c}{(BRISQUE, PI, TE)}\\
          &\multicolumn{1}{c}{(12.3, 4.0, 11.0)}  & \multicolumn{1}{c}{(6.8, 3.9, 11.2)} \\
         \rotatebox{90}{\quad truck} &
         \includegraphics[height=0.23\columnwidth]{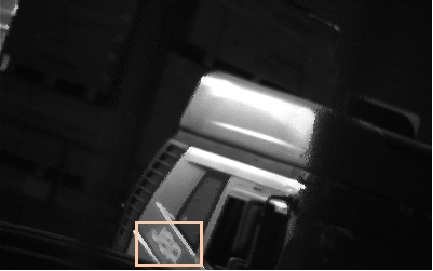} & \includegraphics[height=0.23\columnwidth]{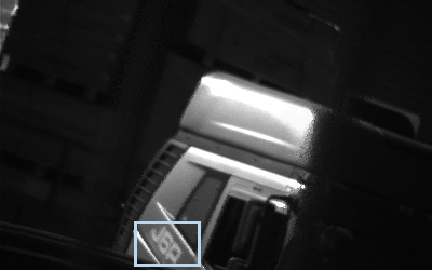} & \includegraphics[height=0.23\columnwidth]{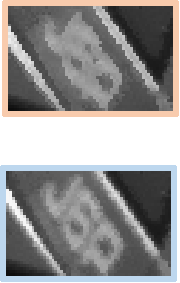}\\
         & \multicolumn{1}{c}{(BRISQUE, PI, TE)} & \multicolumn{1}{c}{(BRISQUE, PI, TE)}\\
         & \multicolumn{1}{c}{(10.2, 5.9, 8.5)} & \multicolumn{1}{c}{(6.1, 5.8, 8.5)} \\
    \end{tabularx}
    \caption{Qualitative comparison of images reconstructed by FSR and SSR.}
    \label{fig:comparison}
\end{figure}

From \tabref{tab:qualitative_metrics}, one may find the quantitative metrics of FSR and SSR are close. This is because, in most cases, the segmentation results of FSR and SSR are the same. However, there are some differences between FSR and SSR. For example, in the "grocery" and "truck" scenes (see \figref{fig:comparison}) captured by the spike camera in a car, the SSR can obtain clearer letters than the FSR. This is because the spike stream segmented by SSR is shorter than or equals to that of FSR, as illustrated in \figref{fig:simplified_workflow}. Such a feature makes SSR more robust to motion blur. To achieve more accurate reconstruction results, SSR requires more calculation resources. It is a trade-off between performance and resource utilization.

\textbf{Reconstruction Speed:} In this section, we compare the reconstruction speed of all the reconstruction methods. We conducted the comparison tests on two platforms: HP EliteDesk 800G4 workstation (referred to PC) and Xilinx VCU118 evaluation board (referred to FPGA). The PC is equipped with Intel® Core™ i7 8700 with Intel (3.2 GHz base frequency, up to 4.6 GHz), 32 GB RAM, and 1080ti 8GB GPU. The FPGA is equipped with Virtex UltraScale+ XCVU9P-L2FLGA2104 (270 Mb URAM, 75.9 Mb BRAM, 1,182k CLB Luts). 

On the PC, we ran all the methods to compare their calculation time. Since some methods do not have the GPU-accelerating version, to be fair, we use the CPU of the PC to do the test. We reconstructed 800 images to obtain the average calculation time of the methods. Correspondingly, the reconstruction speed (FPS) can be obtained by the reciprocal of the calculation time. The calculation time and reconstruction speed are listed in \tabref{tab:PC}. From \tabref{tab:PC}, one can find that the TFI has the highest reconstruction speed achieving 2500 FPS. The reconstruction speed of TFP-32 will decrease about 10 times lower than that of TFI. The reconstruction speed of TFP-64 is about half of that of TFP-32 because its add operations double. The TFSTP is a low-complexity Neural-network-model-based method, whose reconstruction speed is close to that of TFP-64. The reconstruction speeds of our methods are 20.5 FPS and 16.8 FPS, respectively. Compared with these methods (Spk2Img, SSML, and SNM) having similar reconstruction quality, the reconstruction speeds of our methods are greatly improved.  

Besides the PC, we further deployed successfully the FSR and SSR on the FPGA to realize real-time reconstruction (20000FPS). Since the Spk2Img, SSML, SNM, and TFSTP do not have FPGA-implemented versions, we compare the FGPA resource utilizations of TFI, TFP-32, TFP-64, FSR, and SSR, as shown in \tabref{tab:FPGA}. Compared with TFI, TFP-32, and TFP-64, our methods cost more URAM, BRAM, and LUTs. The main reason is that the segmentation process needs to store the $(e_1,e_2)$ and $T_0(i)$, and conducts the zero-order stability judgment. Compared with FSR, the SSR will only increase some LUTs to implement a further step of stability determination on the basis of FSR. Although FSR and SSR require more resources than TFI and TFP, the reconstruction quality is greatly improved. 

To sum up, the FSR and SSR address the motion blur and noise problems of the physical-model-based reconstruction methods and avoid the requirement of large-scale training data of Neural-network-model-based methods. Compared with the reported methods, the FSR and SSR obtain a better tradeoff between reconstruction quality and speed. Due to their parameter-free and easy-implement characteristics, the FSR and SSR can meet the real-time processing requirement of edge-end applications.

\begin{table}[!t]
  \caption{\\Comparison of calculation time and reconstruction speed of different methods on PC}
  \label{tab:PC}
  \centering
  \begin{tabular}{ccc} 
      \hline \hline
      \textbf{Methods} & Calculation Time & Reconstruction Speed\\
      \hline
      TFI \cite{zhu2019retina}          & 0.4ms & 2500 FPS     \\
      TFP-32 \cite{zhu2019retina}     & 3.7ms & 270.0 FPS     \\
      TFP-64 \cite{zhu2019retina}     & 7.5ms & 133.3 FPS    \\
      Spk2Img \cite{zhao2021spk2imgnet}    & 420.0ms & 2.4 FPS     \\
      SSML \cite{chen2022self}         & 232.9ms & 4.3 FPS         \\
      SNM \cite{zhu2023ultra}         & 8340.8ms & 0.1 FPS         \\
      TFSTP \cite{zheng2023capture}   & 7.2ms & 138.9 FPS         \\
      FSR(ours)   & 48.8ms & 20.5 FPS      \\
      SSR(outs)   & 59.5ms & 16.8 FPS      \\
      \hline \hline
  \end{tabular}
\end{table}

\begin{table}[!t]
  \caption{\\Comparison of FPGA Resource utilization of different methods}
  \label{tab:FPGA}
  \centering
  \begin{tabular}{cccccc} 
      \hline \hline
      \textbf{Methods} & URAM & BRAM & LUTs & Reconstruction speed\\
      \hline
      TFI \cite{zhang2022ultra}         &0          &8.0 Mb     &5662       & 20000 FPS     \\
      TFP-32 \cite{zhang2022ultra}     &0          &4.0 Mb     &1905       & 20000 FPS     \\
      TFP-64 \cite{zhang2022ultra}    &0          &8.0 Mb     &4102       & 20000 FPS     \\
      FSR(ours)   &65.7 Mb    &11.5 Mb    &196K       & 20000 FPS      \\
      SSR(outs)   &65.7 Mb    &11.5 Mb    &215K       & 20000 FPS     \\
      \hline \hline
  \end{tabular}
\end{table}

\section{Conclusion}
In this work, we propose a new spike stability theorem, indicating the relationship between spike stream and accumulation rate. With the spike stability theorem, we design two parameter-free reconstruction algorithms called FSR and SSR. We elaborately tested the reconstruction quality and speed of our proposed algorithms and compared them with SOTA algorithms. Experiment results show that, compared with reported algorithms, our algorithms have the best trade-off in image quality and process speed. Further, we demonstrate that the FSR and SSR algorithms can be deployed on the FPGA to realize real-time reconstruction processing, which lays a foundation for edge-end applications of the spike camera. 

In the future, based on the FSR and SSR, we will further study high-level vision tasks, such as object identification, and depth estimation, to realize the real-time vision processing of the spike camera. In addition, as the new spike stability theorem reflects the stable and unstable spike emission rules of the spike camera, it can also be utilized to distinguish the static and motion parts of a scene. So, we will also consider designing high-speed object segmentation and detection algorithms according to this theorem in our future work.

\section*{appendix}

In this section, we give the complete proof of the spike stability theorem. Section III has provided the definition of stability and the relationship between light intensity and stability. To prove the spike stability theorem (Theorem 1), we first introduce two lemmas (\lemref{lem:1} and \lemref{lem:2}). Since the spike stream is also a spike-like stream, the following proof includes the spike-like stream only.

\begin{lemma}
    \label{lem:1}
    Given a spike-like stream $S_0$, if the accumulation rate $Q_0$ is fixed, $S_0$ is a first-order stable stream.
\end{lemma}

\begin{lemma}
    \label{lem:2}
    Given a spike-like stream $S_0$, if the accumulation rate $Q_0$ is fixed, the interval stream of $S_0$, $S_1$, is also a spike-like stream with a fixed accumulation rate $Q_1$.
\end{lemma}

Here we give the proof of \lemref{lem:1}.
\begin{proof}
    \label{prf:1}
    To prove the \lemref{lem:1}, we only need to guarantee the interval stream of $S_0$, $S_1$, is a zero-order stable stream. Given a spike-like stream $S_0$ generated by an integrator $G_0$ with a fixed accumulation rate $Q_0$, the $n^{th}$ readout interval between the $n^{th}$ and $(n+1)^{th} $ firing values in $S_0$, $T_0(n)$, should satisfy

    \begin{align}
        \label{equ:7}
        \begin{cases}
       T_0(n)Q_0\ge M_0^{th},\\
       [T_0(n)-1]Q_0<M_0^{th},\\
         \end{cases}
    \end{align}
    where $M_0^{th}$ is the threshold. \equref{equ:7} ensures the integrator fires the $(n+1)^{th}$ firing value just after $T_0(n)$ checks. According to \equref{equ:7}, it is easy to obtain

    \begin{align}
        \label{equ:8}
       T_0^* \le T_0(n) < 1+T_0^*
    \end{align}
    where
    \begin{align}
        \label{equ:9}
      T_0^* = \frac{M_0^{th}}{Q_0} = [T_0^*]^-+b
    \end{align}
    where $[ ]^-$ is the round down operation, and $b$ is the remainder ($b\in [0, 1)$).
    
    Because the interval $T_0(n)$ is an integer, it can only be $[T_0^*]^-$ or $1+[T_0^*]^-$. Thus, the interval stream $S_1  ({T_0(1), T_0(2), ...})$ can only have at most two elements $[T_0^*]^-+1$ and $[T_0^*]^-$, and the difference between them equals 1. $S_1$ matches the definition of the zero-order stable stream (Definition 2). \lemref{lem:1} is proved.
   
\end{proof}

Next, we give the proof of \lemref{lem:2}.

\begin{proof}
    \label{prf:2_0}
    To prove the \lemref{lem:2}, we need to find an integrator $G_1$ with a threshold $M_1^{th}$ and a fixed accumulation rate $Q_1$ that can generate the stream $S_1 ({T_0(1), T_0(2), ...})$. $T_0(n)$ represents the interval between the $n^{th}$ and $(n+1)^{th} $ firing values in $S_0$, so the status of integrator $G_0$ at the step firing the $(n+1)^{th}$ firing value can be written as 
    
    \begin{equation}
        \label{equ:10}
           {{M}_{0}}\left[ \sum\limits_{i=1}^{n}{{{T}_{0}}\left( i \right)} \right] = {{M}_{0}}\left[ \sum\limits_{i=1}^{n-1}{{{T}_{0}}\left( i \right)} \right]
           +Q_0T_0(n) - M_0^{th}
    \end{equation}
    where ${{M}_{0}}\left[ \sum\limits_{i=1}^{n}{{{T}_{0}}\left( i \right)} \right]$ and ${{M}_{0}}\left[ \sum\limits_{i=1}^{n-1}{{{T}_{0}}\left( i \right)} \right]$ represent respectively the status (before being reset) of $G_0$ when firing the $(n+1)^{th}$ and the $n^{th}$ firing values, $Q_0$ is the accumulation rate, and $M_0^{th}$ is the threshold of $G_0$. We can regard ${{M}_{0}}\left[ \sum\limits_{i=1}^{n}{{{T}_{0}}\left( i \right)} \right]$ as a new $M_1(n)$ that describes the status of $G_1$. Thus, Eq.\eqref{equ:10} can be rewritten as 

    \begin{equation}
        \label{equ:11}
            M_1(n) = M_1(n-1)+Q_0T_0(n) - M_0^{th}
    \end{equation}
    where ${M}_{1}(n)$ and ${M}_{1}(n-1)$ represent the status (before being reset) of $G_1$ at step $n$ and step $n-1$, respectively. 
    Since $T_0(n)$ may be $[T_0^*]^-$ or $[T_0^*]^-+1$, \equref{equ:11} can be expanded as,
        \begin{small}
            \begin{equation}
                \label{equ:12}
                \begin{cases}
                    M_1(n) = & M_1(n-1)+\frac{(1-b)M_0^{th}}{[T_0^*]^-+b}\\
                            & if \quad T(n)=[T_0^*]^-+1 \\
                            \\
                    M_1(n) = & M_1(n-1)+\frac{-bM_0^{th}}{[T_0^*]^-+b} \\
                            & if \quad T(n)=[T_0^*]^-
                \end{cases}
            \end{equation}
        \end{small}
        
        When selecting the $[T_0^*]^-$ as the firing value, let 
         \begin{align}
         \label{equ:13}
            \begin{cases}
                & Q_1 = (1-b)M_0^{th}/\{[T_0^*]^-+b\} \\
                & M_1^{th} = M_0^{th}/\{[T_0^*]^-+b\}
            \end{cases}
        \end{align}
        where $Q_1$ and $M_1^{th}$ can be seen as the fixed accumulation rate and the threshold of $G_0$ respectively. 
        Then, \equref{equ:12} can be rewritten as
        \begin{align}
            \label{equ:14}
            \begin{cases}
                M_1(n) = & M_1(n-1)+Q_1 \\
                         & if \quad T(n)=[T_0^*]^-+1 \\
                M_1(n) = & M_1(n-1)+Q_1-M_1^{th} \\
                         & if \quad T(n)=[T_0^*]^-
            \end{cases}
        \end{align}
         
        \equref{equ:14} formulates the integrate-and-fire process: Given an integrator $G_1$ with threshold $M_1^{th}$, if $M_1(n-1)+Q_1 \ge M_1^{th}$, $S_1$ outputs a firing value $[T_0^*]^-$ and decreases the status of integrator $G_1$ by $M_1^{th}$; If $M_1(n-1)+Q_1 < M_1^{th}$, $S_1$ outputs a resting value $[T_0^*]^-+1$. In other words, we find a $G_1$ with a fixed accumulation $Q_1$ described by \eqref{equ:13} that can generate the stream $S_1 ({T_0(1), T_0(2), ...})$. Of course, it can also select $[T_0^*]^-+1$ as the firing value, and in this case $Q_1=-bM_0^{th}/{\{[T_0^*]+b\}}$ and $M_1^{th}=-M_0^{th}/{\{[T_0^*]+b\}}$. 
        Therefore, $S_1$ is a spike-like stream with a fixed accumulation rate $Q_1$.
\end{proof}

\begin{proof}

    \label{prf:3}
    With \lemref{lem:1} and \lemref{lem:2}, we can easily prove Theorem 1. According to \lemref{lem:1}, given a spike stream $S_0$ with fixed accumulation rate, $S_0$ is a first-stable stream. According to \lemref{lem:2}, the interval stream $S_1$ of $S_0$ is a spike-like stream with fixed accumulation rate. So on and so forth, stream $S_2$, $S_3$,...,$S_n$ will all be spike-like streams as well as first-order stable streams. Therefore, $S_0$ is an absolutely stable stream. 
\end{proof}

{
\bibliographystyle{IEEEtran}
\bibliography{Main}
}

\begin{IEEEbiography}[{\includegraphics[width=1in,height=1.25in,clip,keepaspectratio]{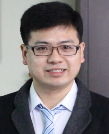}}]{Wei Zhang}
received the B.S. degree in measurement technology and instrumentation and the Ph.D. degree in instrument science and technology from Chongqing University, Chongqing, China, in 2012 and 2016, respectively. From 2017 to 2020, he was a Postdoctoral Fellow with the College of Optoelectronic Engineering, Chongqing University. He is currently an Assistant Researcher with the Department of Networked Intelligence, Peng Cheng Laboratory, Shenzhen, China. His research interests include neuromorphic vision imaging and measurement, optical fiber sensing, lidar, and optimization.
\end{IEEEbiography}

\begin{IEEEbiography}[{\includegraphics[width=1in,height=1.25in,clip,keepaspectratio]{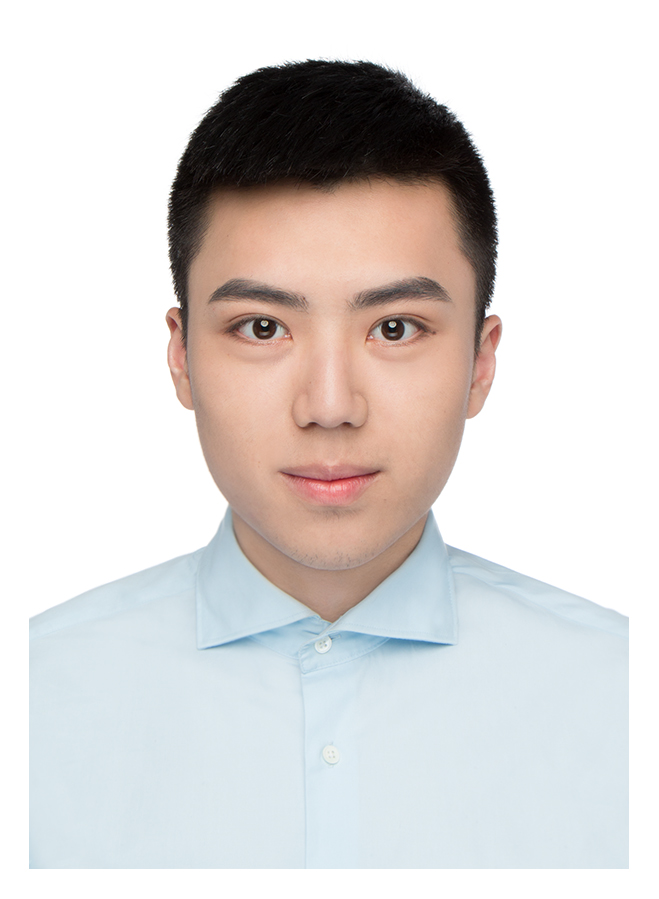}}]{Weiquan Yan}
received the B.Eng. degree in automation from the
Shenzhen University in 2019 and the M.S. degree in intelligent system from the
National University of Singapore in 2021. Currently, he is
working as a deep-learning algorithm engineer at
Peng Cheng laboratory. His research
interests include spike neural networks and few-shot learning.
\end{IEEEbiography}

\begin{IEEEbiography}[{\includegraphics[width=1in,height=1.25in,clip,keepaspectratio]{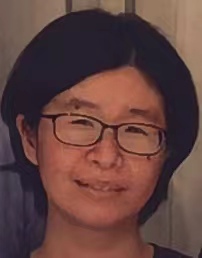}}]{Yun Zhao}
received the B.S. degree in Communication and Information Systems from Harbin Institute of Technology in 2005. She has worked at Huawei Technologies Co., Ltd. for more than 10 years in the FPGA subfield of video conferencing systems. She is currently an engineer in the Network Intelligence Department of Peng Cheng Laboratory, Shenzhen. Her research interest covers FPGA and high-speed video signal processing.
\end{IEEEbiography}

\begin{IEEEbiography}[{\includegraphics[width=1in,height=1.25in,clip,keepaspectratio]{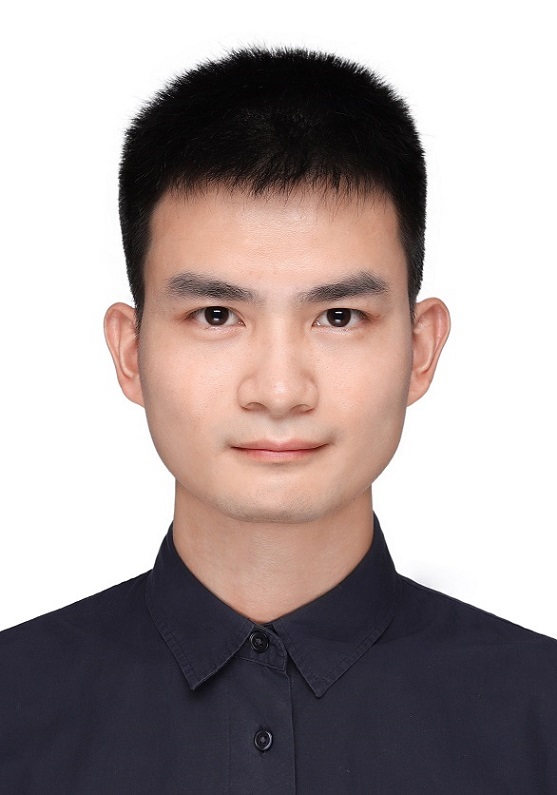}}]{Wenxiang Cheng}
received the B.S. degree in Network Engineering from North University of China in 2018 and the M.S. degree in Computer Technology from Lanzhou University in 2022.Currently he is working as a algorithm engineer at Peng Cheng laboratory. His research interests include deep learning and spike neural networks.
\end{IEEEbiography}

\begin{IEEEbiography}[{\includegraphics[width=1in,height=1.25in,clip,keepaspectratio]{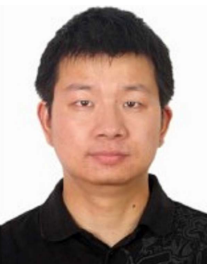}}]{Gang Chen}
received the B.E.
degree in biomedical engineering, the B.S. degree in mathematics and applied mathematics, and the M.S. degree in control science and engineering from Xi’an Jiaotong University, Xi’an, China, in 2008, 2008, and 2011, respectively, and the Ph.D. degree in computer science from the Technical University of Munich, Munich, Germany, in 2016. He is currently an Associate Professor with Sun Yat-Sen University, Guangzhou, China, and Peng
Cheng Laboratory, China. His research interests include deep learning and neural networks.
Dr. Chen received the best paper awards on DATE 2021, ICET 2021, and ESTImedia 2020 and best paper candidate awards on CODES+ISSS 2020. He is an Associate Editor of the Journal of Circuits, Systems and Computers. He also serves as a Reviewer for several Tier-top conferences/journals.
\end{IEEEbiography}

\begin{IEEEbiography}[{\includegraphics[width=1in,height=1.25in,clip,keepaspectratio]{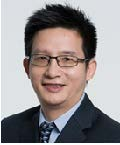}}]{Huihui Zhou}
received his B.E. (1995) in biomedical engineering from Zhejiang University, Hangzhou, China, and his Ph.D.(2000) in physiology from Chinese Academy of Medical Sciences \& Peking Union Medical College, Beijing, China Currently, he is a Professor in Peng Cheng Laboratory, Shenzhen, China. His research interests include neural circuits of visual cognition, brain-machine interface, and spiking neural networks.
\end{IEEEbiography}

\begin{IEEEbiography}[{\includegraphics[width=1in,height=1.25in,clip,keepaspectratio]{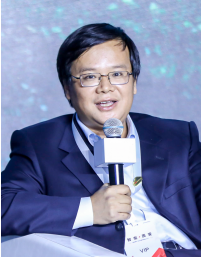}}]{Yonghong Tian}
is currently the Dean of School of Electronics and Computer Engineering, a Boya Distinguished Professor with the School of Computer Science, Peking University, China, and is also the deputy director of Artificial Intelligence Research, Peng Cheng Laboratory, Shenzhen, China. His research interests include neuromorphic vision, distributed machine learning, and multimedia big data. He is the author or coauthor of over 300 technical articles in refereed journals and conferences. He was the recipient of the Chinese National Science Foundation for Distinguished Young Scholars in 2018, two National Science and Technology Awards, and three ministerial-level awards in China. He is a Fellow of IEEE, a senior member of CIE and CCF, a member of ACM.
\end{IEEEbiography}
\end{document}